\documentclass{article}

    \usepackage[preprint]{neurips_2022}

\usepackage[utf8]{inputenc} 
\usepackage[T1]{fontenc}    
\usepackage[colorlinks = true,
            linkcolor = blue,
            urlcolor  = dkgreen,
            citecolor = blue,
            anchorcolor = blue]{hyperref}
\usepackage{url}            
\usepackage{booktabs}       
\usepackage{amsfonts}       
\usepackage{nicefrac}       
\usepackage{microtype}      
\usepackage{xcolor}         
\usepackage{graphicx}
\usepackage{subfigure}
\usepackage{booktabs}
\usepackage{caption}
\usepackage{multirow}
\usepackage{amsmath}
\usepackage{amssymb}
\usepackage{mathtools}
\usepackage{amsthm}
\usepackage{caption}
\usepackage{changepage}
\usepackage{tabularx}

\usepackage{listings}
\usepackage{color}
\usepackage{cleveref}

\definecolor{dkgreen}{rgb}{0,0.6,0}
\definecolor{gray}{rgb}{0.5,0.5,0.5}
\definecolor{mauve}{rgb}{0.58,0,0.82}

\title{Patch-based Object-centric Transformers for \\Efficient Video Generation}

\author{%
  Wilson~Yan$^{1}$ \And
  Ryo Okumura$^{2}$ \And
  Stephen James$^{1}$ \And
  Pieter Abbeel$^{1}$
  \thanks{$^{1}$ UC Berkeley $^{2}$ Panasonic. Correspondance to wilson1.yan@berkeley.edu}
}

\makeatletter
\def\thanks#1{\protected@xdef\@thanks{\@thanks
        \protect\footnotetext{#1}}}
\makeatother
\begin{document}

\maketitle

\begin{abstract}
In this work, we present Patch-based Object-centric Video Transformer (POVT), a novel region-based video generation architecture that leverages object-centric information to efficiently model temporal dynamics in videos. We build upon prior work in video prediction via an autoregressive transformer over the discrete latent space of compressed videos, with an added modification to model object-centric information via bounding boxes. Due to better compressibility of object-centric representations, we can improve training efficiency  by allowing the model to only access object information for longer horizon temporal information. When evaluated on various difficult object-centric datasets, our method achieves better or equal performance to other video generation models, while remaining computationally more efficient and scalable. In addition, we show that our method is able to perform object-centric controllability through bounding box manipulation, which may aid downstream tasks such as video editing, or visual planning. Samples are available at \href{https://sites.google.com/view/povt-public}{https://sites.google.com/view/povt-public}
\end{abstract}

\begin{figure*}[!ht]
    \centering
    \includegraphics[width=\textwidth]{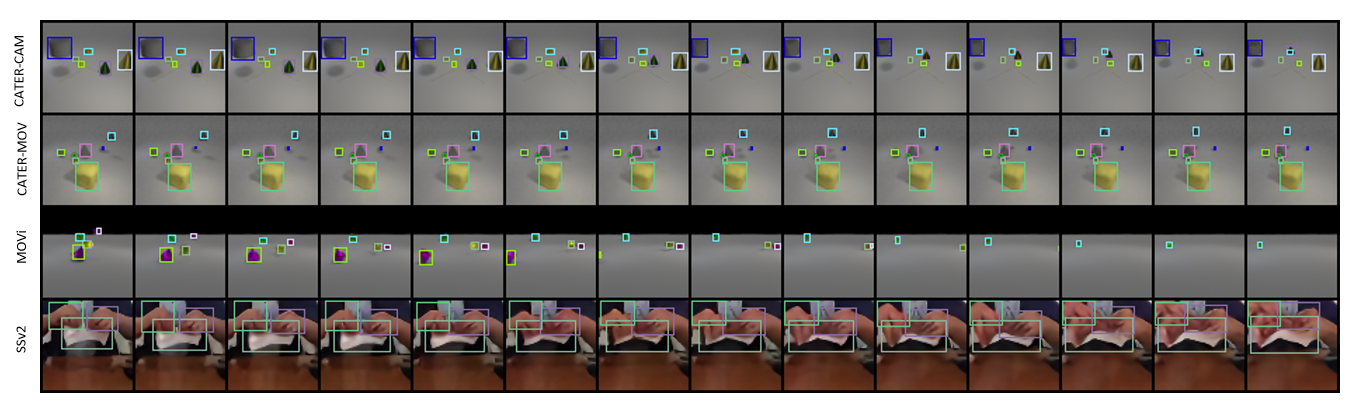}
    \caption{Video prediction samples along with predicted bounding boxes}
    \label{fig:povt_bbox}
\end{figure*}
\section{Introduction}
In recent years, generative modeling has seen a large amount of success in various domains, including images~\citep{kingma2013auto,goodfellow2014generative,kingma2018glow,dinh2016density}, audio~\citep{oord2016wavenet,prenger2019waveglow,oord2017parallel}, and language~\citep{radford2019language,brown2020language}. Rapid progress in these domains can be largely attributed to deep learning methods through a combination of algorithmic design and the wider availability of more compute resources.

Although strong progress has been made in high-fidelity image generation~\citep{oord2016pixel,brock2018large,karras2019style,ho2020denoising}, videos remain a difficult modality to train, primarily due to the complexity of the data distribution as well as the high compute requirements for modeling larger temporal or spatial resolution videos. As such, most works in high fidelity video generation have been limited to short sequences of 1-2 seconds at a resolution of $64\times 64$~\citep{clark2019adversarial,yan2021videogpt}, with a few exceptions that aim to scale to higher spatial, temporal resolutions, or both~\citep{anonymous2022autoregressive,skorokhodov2021stylegan,tian2021good}. 

In this work, we focus on video generation with autoregressive likelihood-based models, since they are well understood, easy to train, and do not suffer from posterior collapse or mode collapse. However, autoregressive modeling in pixel space is prohibitively expensive~\citep{weissenborn2019scaling}. As such, recent work seeks to address this issue by learning autoregressive priors on compressed latents from a discrete autoencoder~\citep{yan2021videogpt,rakhimov2020latent,le2021ccvs,walker2021predicting}. Doing so brings about orders of magnitudes of savings in computational cost compared to modeling pixels directly.

Our method, Patch-based Object-centric Video Transformer (POVT), proposes to use object-centric information (via bounding boxes) as a means to acquire more compressible representations for video sequences, resulting in improved computational efficiency. In addition, object-centric representations encode a high-level compositional hierarchy that may benefit models for better generalization and robustness to new tasks. However, in order to generalize to complex video, we cannot solely rely on local object-centric representations, but also must consider the wider global context; for example, generating scenes with dynamic backgrounds, or moving non-object entities, such as changing lighting, camera movement, or flowing water. As such, we seek to design an architecture that predicts object dynamics at the top level of a hierarchy, and then uses this information for more informed prediction on the whole scene image. 

We build upon prior work in video generation architectures using autoregressive discrete latent variable models that combine VQ-VAE~\citep{oord2017neural} and transformers~\citep{vaswani2017attention,brown2020language} by introducing a novel two-stream attention module that jointly models VQ-VAE discrete image latents along with predicting object bounding boxes for each frame. We show that using object-centric representations constructed from bounding box information allows for more efficient and scalable video prediction by extending temporal context only along the object-stream.

In summary, we present the following key contributions in this paper:
\begin{itemize}
    \item To the best of our knowledge, we propose the first object-centric video prediction model able to work on more complex in-the-wild video data
    
    \item We demonstrate that POVT achieves comparable or better predictive accuracy compared to similar models with more than 2x savings in computational cost, and even better scaling on longer horizon videos (due to quadratic complexity of attention in sequence length)
    
    \item We show that better computational efficiency allows us to scale to larger models on more complex datasets to achieve higher fidelity video prediction quality
    
    \item Lastly, we show that POVT allows for controllability and manipulation of objects in videos
\end{itemize}

\section{Related Work}
\subsection{Video Synthesis and Prediction}
Video synthesis and prediction are two closely related areas, where prediction can be commonly viewed as conditional video synthesis. Prior work in video generation (prediction or synthesis) can generally be divided into GAN-based methods and likelihood-based methods.

GAN~\citep{goodfellow2014generative} video generation models are the most commonly proposed methods, primarily centered around innovation through architecture design. Prior work such as VGAN~\citep{vondrick2016generating}, MoCoGAN~\citep{tulyakov2018mocogan}, and TGAN~\citep{saito2017temporal} leverage the notion of separate content and motion as an inductive bias for encouraging smoother video generation. MoCoGAN-HD~\citep{tian2021good} extends to larger resolution video by learning to traverse the latent space of pretrained image GANs such as StyleGANv2~\citep{karras2019style}. TGANv2~\citep{saito2018tganv2} proposes more efficient GAN training through a hierarchical generation mechanism that allows for training on sparse subsets of video sequences. DVD-GAN~\citep{clark2019adversarial} and TrIVD-GAN~\citep{luc2020transformation} investigate more efficient discriminator designs. More recently, methods such as StyleGAN-V~\citep{skorokhodov2021stylegan} and DI-GAN~\citep{yu2022generating} adopt continuous interpretations of video modeling and are able to generate long spatio-temporal resolution videos through training on very sparse videos. However, GAN-based video models still suffer from mode collapse and have trouble generating diverse outputs on complex datasets.

Another broad class of video generation models are likelihood-based methods, such as variational and autoregressive models. \citet{villegas2019high} and \citet{castrejon2019improved} extend to videos through optimizing an ELBO consisting of a temporal decomposition of latents. \citet{lee2018stochastic} adds an adversarial loss to encourage more realistic generation. FitVid~\citep{babaeizadeh2021fitvid} introduces a more efficient, high capacity architecture. In contrast to optimizing an ELBO, autoregressive methods model log-likelihood. Video Pixel Networks~\citep{kalchbrenner2017video} learns a PixelCNN~\citep{oord2016conditional} decoder conditioned on frame history encoded using an LSTM~\citep{hochreiter1997long}. Subscale Video Transformer~\citep{weissenborn2019scaling} defines a unique spatio-temporal subscale pixel ordering to model videos. Due to these models typically having long sampling times and high compute requirements, more recent works~\citep{rakhimov2020latent,yan2021videogpt,walker2021predicting,le2021ccvs,anonymous2022autoregressive} have investigated using autoregressive models on the latent space output of a discrete encoder, such as VQ-VAE~\citep{oord2017neural} or VQ-GAN~\citep{esser2021taming}. This allows for training higher capacity models on less compute, as well as faster inference time; however, there is an inherent trade-off due to the lossy compression of discrete autoencoders. 

In this work, we show that object-centric representations, through the use of bounding boxes, allows for efficient and scalable video generation. While hierarchical representations, such as keypoints~\citep{villegas2017learning,minderer2019unsupervised,kim2019unsupervised} and segmentation maps~\citep{luc2017predicting,luc2018predicting,lee2021revisiting} have been shown to improve long-horizon video generation, our work is the first to explore a bounding box representation for video generation which extends to realistic video.


\subsection{Object-centric Generative Models}
There is a substantial amount of literature on object-centric representation learning and generative modeling, in which objects can be discovered in an unsupervised manner. Image-based models such as MONet~\citep{burgess2019monet}, GENESIS~\citep{engelcke2019genesis}, SPACE~\citep{lin2020space}, and IODINE~\citep{greff2019multi} bias their models to encourage object-centric decomposition. Sequential object-centric models such as SQAIR~\citep{kosiorek2018sequential}, R-SQAIR~\citep{stanic2019r}, SCALOR~\citep{jiang2019scalor}, STOVE~\citep{kossen2019structured}, and G-SWM~\citep{lin2020improving} leverage the use of videos to track and identify objects in a scene. Slot Attention~\citep{locatello2020object} uses iterative attention on several fixed or randomly generated slots to identify individual objects. SAVi~\citep{kipf2021conditional} extends on Slot Attention to videos by propagating and updating slots over time. AG2Vid~\citep{bar2020compositional} synthesizes videos given action graphs. OCVT~\citep{wu2021generative} uses a transformer over object-centric representations derive from a pre-trained SPACE model to predict videos. However, unlike our method, it does not account for the global context of the scene, and is therefore unable to generalize to diverse and dynamic backgrounds. In addition, it is unable to model stochastic dynamics, a crucial component for any video prediction model.

\begin{figure*}[!ht]
    \centering
    \includegraphics[width=0.9\textwidth]{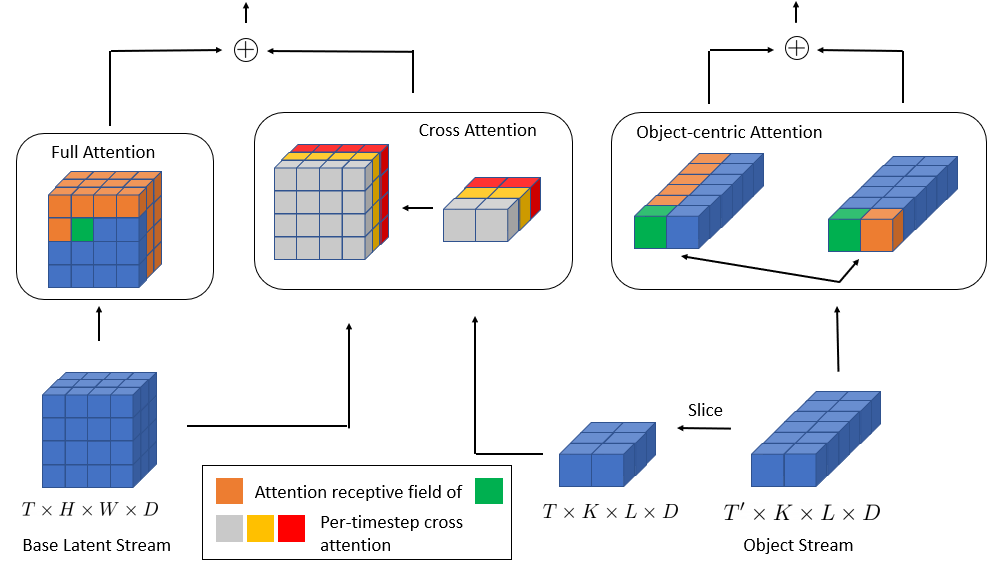}
    \caption{Object-centric attention block for POVT. A base latent stream and object stream are inputted, with a combination of self-attention on the base latent stream, object-centric attention on the object stream, and cross-attention for inter-stream communication. Each object (max $K$) is represented by an $L\times D$ representation, which is a concatenation of bounding box and patch encoding information}
    \label{fig:arch1}
\end{figure*}

\section{Method}
In this section, we describe POVT, an extension on existing autoregressive discrete latent variable models to incorporate object-centric information for efficient and scalable video generation.

We first learn a VQ-VAE~\citep{oord2017neural} encoder to map individual video frames to a 2D sequence of discrete latents. In order to incorporate object-centric information, we consider the bounding boxes of a given set of tracked objects in the scene.  Let the VQ-VAE discrete latents for a video sequence of length $T$ be denoted as $z_1, \dots, z_T$, and the bounding boxes for $K$ objects be $\{(b^{1}_1,\dots,b^{1}_T), \dots, (b^{K}_1,\dots,b^{K}_T)\}$, where we consider object tracks as a set and thus permutation invariant. Each bounding box $b^{i}_t$ is represented as a tuple $(o_{pres}, o_{where}, o_{scale})$. $o_{pres}$ is a binary random variable defining the presence of the object in the scene, generally changed through occlusion or movement in or out of a scene. $o_{where}$ and $o_{scale}$ are only set when $o_{pres} = 1$, and consist of the location and scale of the bounding box parameters of a given object. Bounding box parameters are normalized to $[0, 1]$ and quantized uniformly over $64$ bins (the resolution we train on).

Following VQ-VAE training, we train a transformer to model the joint distribution over discrete codes and bounding boxes $p(b_1,z_1, \dots, b_T, z_T)$ autoregressively  with the following factorization:
$$\prod_{t=1}^T \left(\prod_{k=1}^K p(b^{k}_t | b^{<k}_t, b_{<t}, z_{<t})\right)p(z_t | b_{\leq t}, z_{<t})$$
Each $b^{k}$ is modeled autoregressively in the order $o_{pres}, o_{where}, o_{scale}$, and similarly for each $z_t$ in 2D raster scan order. During inference, the model alternates between modeling bounding box dynamics, and generating the corresponding set of discrete latents conditioned on the predicted bounding boxes for each object.

In order to allow for more computationally efficient training, we introduce an efficient object-centric attention architecture (Section \ref{object_centric_transformer}) and a training regime in which detailed object-centric information, along with coarse global information, is extended back in time (Section \ref{training}).



\begin{figure*}
            \centering
    \includegraphics[width=0.9\textwidth]{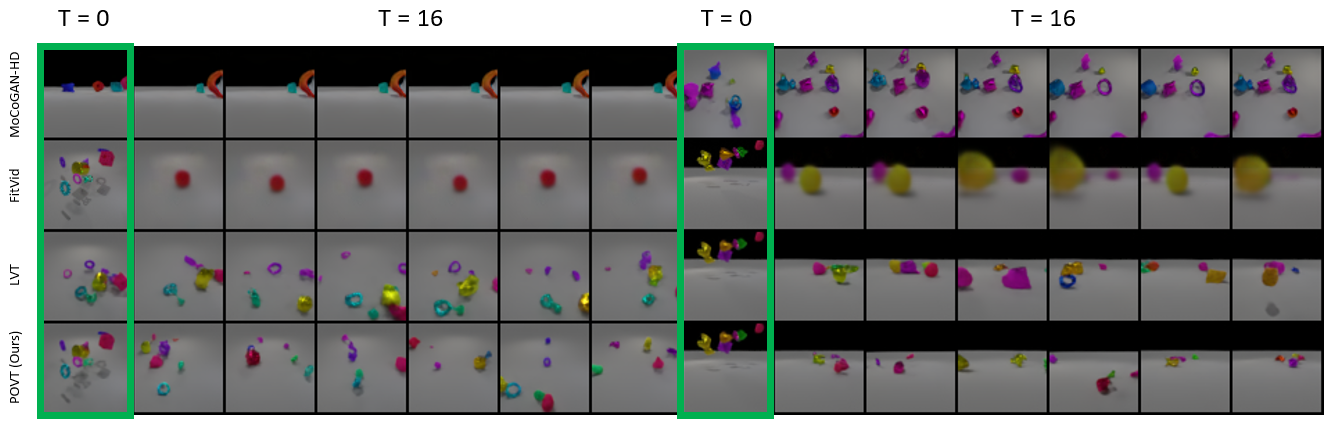}
    \caption{Diversity of video prediction results for each model. For each model, we condition on 1 frame and visualize the final frame of 6 different predicted futures. For MoCoGAN-HD, we generate multiple futures starting with 1 randomly sampled latent vector. MoCoGAN-HD and FitVid generally show low sample diversity compared to LVT and POVT}
    \label{fig:diversity}
\end{figure*}

\subsection{Object-centric Transformer}
\label{object_centric_transformer}
We introduce a two-stream attention architecture for modeling discrete latents and object dynamics separately, with carefully added inter-stream attention operations. 

\textbf{Constructing object representations}\\
We use a Spatial Transformer~\citep{jaderberg2015spatial} to extract region patches from each frame for every bounding box, where each region patch is resized to a fixed resolution. In the case where a bounding box is invalid with $o_{pres} = 0$, we set bounding box latents and patches to zero. Each patch is then downsampled using a strided convolution to a fixed resolution, similar to ViT~\citep{dosovitskiy2020image}. When training, we shift and pad all patch representations one timestep forward since bounding boxes at a timestep $t$ should not be able to see their corresponding patch representations, as they can only be extracted once the full image has been generated. For $t = 0$, we pad with a learned patch token that is broadcasted over all objects at that timestep. Patch representations are then concatenated with their corresponding discretized bounding boxes.

\textbf{Object-centric attention}\\
Figure~\ref{fig:arch1} shows the proposed architecture of one attention block, and pseudo code is provided in Appendix~\ref{appendix:code}. Attention over objects is decomposed into: (1) attention over different objects of the same timesteps, and (2) attention over same objects of different timesteps. Intuitively, the first step models object interactions at each timestep, and the second step summarizes the temporal evolution of an object's history.

Attention over the base latent stream is full self-attention over all discrete latents. We incorporate object-centric information into the discrete latents through full cross-attention only with objects at the current timesteps. Specifically, $z_t$ only attends to $b_t$ and its corresponding patch representations, and does not attend to the full object history. This architecture is far more efficient than attending over the entire video representation; increasing computational efficiency.

Our transformer architecture follows GPT~\citep{radford2019language} with each attention block replaced with our object-centric attention block. We use spatio-temporally broadcasted positional embeddings~\citep{yan2021videogpt} for the base latent stream. Similarly, object-temporally broadcasted positional embeddings are applied to the encoded objected bounding boxes and object region encodings.

\subsection{Combining Global and Local Representations for Efficient Training}
\label{training}

In-scene dynamics are most relevant to the object representations; therefore, a significant reduction in computational costs can be achieved by only extending object representations back in time, rather than the full set of discrete latents. However, we still require a few frames of discrete latents in order to model the global, non-object dynamics, such as camera jitter or changing backgrounds. We can more efficiently model video sequences of length $T$ by defining $T'$ such that we model the distribution:
$$p(b_{T-T':T}, z_{T-T':T} | b_{T-T'-\tau:T-T'-1})$$
where $\tau$ is uniformly sampled every iteration from $[0, T - T']$. $\tau$ is sampled in order to train on the full range of unconditional and conditional distributions needed during inference. This allows us to incorporate longer horizon temporal information about the scene in a more computationally efficient manner due to the higher compressibility of object latents compared to the global discrete latents from the VQ-VAE. In experiments, $T$ is set to the same horizons as baseline models, and $T'$ set depending on the dataset, where datasets with more non-object dynamics require greater $T'$ to include more full-frame temporal information.


\begin{table*}[t!]
\centering
\captionsetup{width=\textwidth}
\caption{Results for video prediction on CATER-CAM, CATER-MOV, and MOVi, comparing our method against baseline methods FitVid and LVT. Our method is competitive with a fraction of the training and computational cost. Best performing models for each metric are in bold, with multiple highlighted if within a 95\% confidence interval}
\begin{tabular}{@{}lcccccccc@{}}
\toprule
\textbf{CATER-CAM}               & FVD$\downarrow$ & PSNR$\uparrow$ & SSIM$\uparrow$ & LPIPS$\downarrow$ & Params       & GFLOP       & GPU Days   \\ \midrule
FitVid               & 275             & 25.2           & 0.819          & 0.106             & 78M          &       239      & 1.5          \\
LVT                  & \textbf{219}    & \textbf{26.4}  & \textbf{0.864} & \textbf{0.0662}   & \textbf{26M}          & 181         & 2          \\
POVT (ours) & \textbf{223}    & 25.6  & \textbf{0.857} & 0.0701   & \textbf{26M} & \textbf{81} & \textbf{1} \\ \bottomrule
\end{tabular}
\begin{tabular}{@{}lcccccccc@{}}
\toprule
\textbf{CATER-MOV}               & FVD$\downarrow$ & PSNR$\uparrow$ & SSIM$\uparrow$ & LPIPS$\downarrow$ & Params & GFLOP & GPU Days \\ \midrule
FitVid               & 181             & \textbf{30.4}           & 0.87           & 0.0971             & 78M    &   239    & 1.5        \\
LVT                  & \textbf{166}             & 30.2           & \textbf{0.903}          & \textbf{0.0605}            & \textbf{26M}    & 181   & 2        \\
POVT (ours) & \textbf{164}             & 30.1           & \textbf{0.902}          & \textbf{0.0584}            & \textbf{26M}    & \textbf{81}    & \textbf{1}        \\ \bottomrule
\end{tabular}
\begin{tabular}{@{}lcccccccc@{}}
\toprule
\textbf{MOVi}               & FVD$\downarrow$ & PSNR$\uparrow$ & SSIM$\uparrow$ & LPIPS$\downarrow$ & Params & GFLOP & GPU Days \\ \midrule
FitVid               & 983            & 22.5           & 0.807           & 0.313              & 78M    &    239   & 1.5        \\
LVT                  & \textbf{222}             & 22.6           & \textbf{0.809}          & \textbf{0.163}             & \textbf{26M}    & 181   & 2        \\
POVT (ours) & \textbf{226}             & \textbf{22.9}           & \textbf{0.815}          & \textbf{0.157}             & \textbf{26M}    & \textbf{81}    & \textbf{1}       \\ \bottomrule
\end{tabular}
\label{Table:cater_movi_pred}
\end{table*}

\begin{table*}[t!]
\centering
\caption{Quantitative evaluation of single-frame video prediction on Something-Something V2. Our method scales better while retaining access to longer temporal context. All methods perform similarly in terms of PSNR, SSIM, and LPIPS with little difference compared to the baseline of repeating the initial ground truth frame. We hypothesize that this is due to the highly stochastic nature of SSv2. As such, we believe FVD is a more reliable metric to measure the realism of predicted dynamic motion.}
\label{table:ss_pred}
\begin{tabular}{@{}lcccccccc@{}}
\toprule
\textbf{SSv2}               & FVD$\downarrow$ & PSNR$\uparrow$ & SSIM$\uparrow$ & LPIPS$\downarrow$ & Params & GFLOP & GPU Days \\ \midrule
Repeat Frame & 600 & 19.0 & 0.611 & 0.172 & - & - & - & - \\
FitVid               & 1028             & 19.5           & 0.618           & 0.241             & 315M    &   930    & 16h        \\
LVT-S (8 frames)                  & 558             & 19.1           & 0.615          & 0.180            & 105M    & 551   & 16        \\
LVT-L (4 frames)                  & 530             & 19.2           & 0.624          & 0.173            & 206M    & 482   & 16        \\
POVT (ours) & \textbf{497}             & 19.1           & 0.622          & 0.170            & 206M    & 602    & 16        \\ \bottomrule
\end{tabular}
\end{table*}

\section{Experiments}
In the following section, we evaluate our method to answer the following questions:
\begin{itemize}
    \item Can we learn more efficient video generation models with object-centric information?
    \item Can we learn object-centric video models that are able to model real video?
    \item Can we scale to larger models for generating higher-fidelity video?
\end{itemize}
\subsection{Datasets}
In this section, we briefly introduce each dataset we evaluate our method on. More details on dataset generation and processing can be found in Appendix~\ref{appendix:dataset}.\\\\
\textbf{CATER}: CATER~\citep{girdhar2019cater} consists of a set of simple objects (cubes, spheres, cones) that randomly move around in a scene. The original dataset does not contain any segmentation or bounding box information for objects. Therefore, we use a version of CATER with object segmentations introduced in~\citet{kabra2021simone} (that we call CATER-CAM), which contains 56,464 videos of 33 frames with random camera movement, and up to 10 objects, where a maximum of 2 objects move at a time. In addition, we generate another more complex version of CATER (called CATER-MOV) with 10,000 videos of 100 frames, with a maximum of 10 objects, and at most 10 objects moving around. 

\textbf{MOVi}: Multi-Object Video (MOVi) is a dataset introduced in Kubric~\citep{greff2021kubric} which simulates 3D rigid body interactions in a Blender environment. Scenes generally consist of a subset of randomly generated objects that fall with randomly sampled force vectors. We simulate 10,000 scenes of 5-10 objects and render videos of 24 frames. 

\textbf{Something-Something V2 (SSv2)}: SSv2~\citep{goyal2017something} is a real-world, action recognition dataset with a complex suite of action classes that generally includes object interaction. `Something Else'~\citep{materzynska2020something} provides bounding box annotations for a large subset of the videos, computed using a pretrained Faster R-CNN object detector. Following \citet{materzynska2020something}, we use a simple online tracker to group bounding boxes into separate tracks, and restrict training to a subset of 150,000 videos with a maximum of $4$ objects.  

\begin{table}[]
\begin{adjustwidth}{-1.75cm}{-1.75cm}
\begin{tabularx}{\linewidth}{*{2}{>{\centering\arraybackslash}X}}
\caption{Unconditional video synthesis on CATER-CAM}
\label{Table:cater_cam_synth}
\begin{tabular}{@{}lcccc@{}}
\toprule
\textbf{CATER-CAM}               & FVD$\downarrow$ & Param & GFLOP & GPU Days   \\ \midrule
MoCoGAN-HD           & 294             & 72M   &  242     & 2 \\
POVT (ours) & \textbf{273}             & \textbf{26M}   & \textbf{81}    & \textbf{1} \\ \bottomrule
\end{tabular}

&
\caption{Unconditional video synthesis on CATER-MOV}
\label{Table:cater_mov_synth}
\begin{tabular}{@{}lcccc@{}}
\toprule
\textbf{CATER-MOV}               & FVD$\downarrow$ & Param & GFLOP & GPU Days   \\ \midrule
MoCoGAN-HD           & 236             & 72M   &  242     & 2 \\
POVT (ours) & \textbf{193}             & \textbf{26M}   & \textbf{81}    & \textbf{1} \\ \bottomrule
\end{tabular}

\\

\caption{Unconditional video synthesis on MOVi}
\label{Table:movi_synth}
\begin{tabular}{@{}lcccc@{}}
\toprule
\textbf{MOVi}               & FVD$\downarrow$ & Param & GFLOP & GPU Days   \\ \midrule
MoCoGAN-HD           & \textbf{282}             & 72M   &   242    & 2 \\
POVT (ours) & 295             & \textbf{26M}   & \textbf{81}    & \textbf{1} \\ \bottomrule
\end{tabular}
&
\caption{Unconditional video synthesis on SSv2}
\label{table:ss_synth}
\begin{tabular}{@{}lcccc@{}}
\toprule
\textbf{SSv2}               & FVD$\downarrow$ & Param & GFLOP & GPU Days   \\ \midrule
MoCoGAN-HD           & \textbf{420}             & \textbf{179M}   &  \textbf{542}     & 16 \\
POVT (ours) & 562             & 206M   & 602    & 16 \\ \bottomrule
\end{tabular}

\end{tabularx}
\end{adjustwidth}
\end{table}


\subsection{Baselines}
We compare our method against a baseline from each common class of video generation models. To the best of our knowledge, there do not exist any video prediction methods that use bounding box information (ground-truth or from a pretrained object detector), so we compare against unconditional baselines, and show that our method not only allows for a more efficient models, but also is able to extend to real video unlike existing object-centric video models. We do not compare against OCVT~\citep{wu2021generative} since it would perform poorly on our datasets as OCVT is unable to model stochastic dynamics and is limited to deterministic settings.
\begin{itemize}
    \item \textbf{FitVid}~\citep{babaeizadeh2021fitvid} is a variational video prediction method that outperforms popular baselines such as SVG~\citep{denton2018stochastic}, SAVP~\citep{lee2018stochastic}, SV2P~\citep{babaeizadeh2017stochastic}, and GHVAE~\citep{wu2021greedy}. We choose FitVid over object-centric variational approaches such as G-SWM~\citep{lin2020improving} since they exhibited worse performance than FitVid, and have difficulty generalizing to realistic video such as SSv2.
    \item \textbf{MoCoGAN-HD}~\citep{tian2021good} learns a video generation model by traversing the latent space of a pretrained image generator. It is able to produce high fidelity video generation with minimal computational cost compared to models such as DVD-GAN~\citep{clark2019adversarial} or TrIVD-GAN~\citep{luc2020transformation}.
    \item \textbf{Latent Video Transformer (LVT)}~\citep{rakhimov2020latent} is an AR model that learns a transformer over discrete latent codes of videos encoded using a VQ-VAE. Other baseline models primarily propose modifications to learn better discrete codes, such as through flow-conditional reconstruction~\citep{le2021ccvs}, using a VQ-GAN~\citep{anonymous2022autoregressive}, or using a 3D VQ-VAE~\citep{yan2021videogpt}, which we believe are orthogonal to our approach which focuses on the transformer architecture. Therefore, we opt for the simplest baseline to adapt our model to. Both LVT and our method POVT use the same VQ-VAE on each dataset.
\end{itemize}

\subsection{Training Details}
For all datasets, we train on videos of 8 frames (16 frames did not increase performance) scaled to $[0, 1]$, and evaluate on generated videos of 16 frames. 
For CATER-CAM, CATER-MOV, and MOVi, we condition on 1 full frame, and 7 timesteps of object representations. Full frame conditioning is to account for random camera and lighting changes per video. For SSv2, we condition on 3 full frames and 7 timesteps of object representations to model local non-object dynamics of the complex videos. All baseline methods are trained on the full $8$ frames. More details on hyperparameters for POVT and baselines can be found in Appendix~\ref{appendix:arch_hyperparams}. All experiments are trained on a maximum of 8 RTX A5000 GPUs each with 24GB of memory.

\begin{figure*}[t!]
    \centering
    \includegraphics[width=0.9\textwidth]{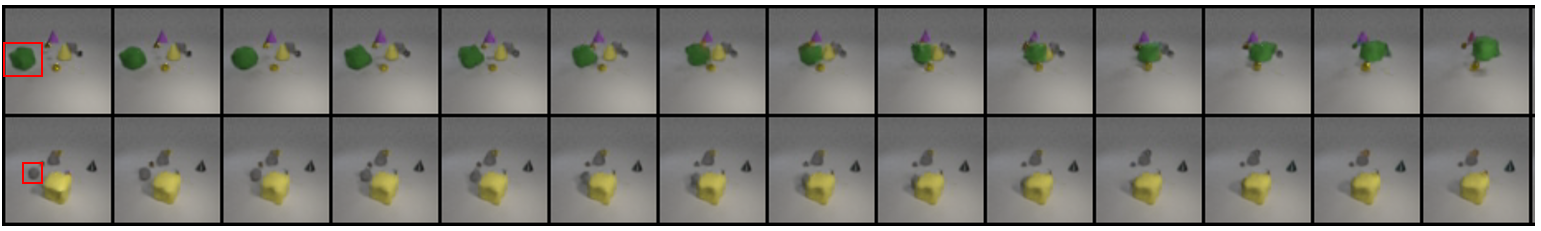}
    \caption{One object (in red) is translated to a random pixel. Note that occlusion is maintained during translation in both cases where the object occludes (top) and is occluded by other objects (bottom)}
    \label{fig:contr_translate}
\end{figure*}

\begin{table*}[]
\centering
\captionsetup{width=\textwidth}
\caption{Ablation on the number of full frames to condition on. All experiments of POVT condition on 7 timesteps of object history in addition to the specified number of full frames}
\label{table:abl_cond}
\begin{tabular}{@{}ccccccc@{}}
\toprule
Method                                & \multicolumn{1}{l}{\# Cond Frames} & FVD$\downarrow$ & \multicolumn{1}{l}{PSNR$\uparrow$} & \multicolumn{1}{l}{SSIM$\uparrow$} & \multicolumn{1}{l}{LPIPS$\downarrow$} & GFLOP \\ \midrule
\multirow{4}{*}{LVT}                  & 1                                  & 218             & 28.4                               & 0.884                              & 0.0838                                & 38    \\
                                      & 2                                  & 207             & 29.0                               & 0.890                              & 0.0751                                & 56    \\
                                      & 4                                  & 165             & 29.6                               & 0.895                              & 0.0646                                & 77  
                                      \\
                                      & 7                                  & 166             & 30.2                               & 0.903                              & 0.0605                                & 181  
                                      \\ \midrule
\multirow{3}{*}{POVT (ours)} & 1                                  & 164             & 30.1                               & 0.902                              & 0.0583                                & 81    \\
                                      & 2                                  & 158             & 30.0                               & 0.897                              & 0.0609                                & 100   \\
                                      & 4                                  & 160             & 30.1                               & 0.898                              & 0.0599                                & 121   \\ \bottomrule
\end{tabular}
\end{table*}

\subsection{Evaluation}

\textbf{Metrics:} For video prediction tasks, we use Fr\'echet Video Distance (FVD)~\citep{unterthiner2019fvd}, PSNR~\citep{huynh2008scope}, SSIM~\citep{wang2004image}, and LPIPS~\citep{zhang2018unreasonable}. In order to account for stochasticity of video prediction, we follow prior work~\citep{babaeizadeh2017stochastic,villegas2019high,babaeizadeh2021fitvid} and take the best matching trajectory with ground truth out of 100 samples. For final evaluation, we take the average over $128$ test examples. For video synthesis, we evaluate only on FVD. We compute FVD over batches of $256$ and average over 16 runs. Qualitative samples can be found on our paper website at \href{https://sites.google.com/view/povt-public}{https://sites.google.com/view/povt-public}.

\textbf{CATER-CAM / CATER-MOV / MOVi:} \Cref{Table:cater_movi_pred,Table:cater_cam_synth,Table:cater_mov_synth,Table:movi_synth} show video prediction and synthesis results on CATER-CAM, CATER-MOV, and MOVi datasets. For CATER-MOV and MOVi, our method performs competitively, or better than baseline methods at a fraction of the computation cost in terms of GLOPs and GPU-days. Our method slightly underperforms LVT for CATER-CAM, which we hypothesize may be due to constant shifts in scale of object representations due to zooming camera movement. Sample-wise, each model shows similar behavior across all object datasets (CATER-CAM, CATER-MOV, and MOVi). MoCoGAN-HD produces high quality images, but has difficult retaining object consistency across time, where objects would constantly disappear, appear, and morph to other objects. FitVid learns better reconstruction than LVT and POVT, but samples often result in blurry movement. In addition, FitVid and MoCoGAN-HD generally have difficulty modeling diversity of future predictions and generate predictions with very low diversity compared to our method, shown in Figure~\ref{fig:diversity}.

\textbf{Something-Something V2:} Tables~\ref{table:ss_pred} and~\ref{table:ss_synth} show quantitative results on SSv2. For consistency, we report PSNR, SSIM, and LPIPS. However, metrics across all models show little difference compared to a naive baseline of repeating the first frame over the entire trajectory (Repeat Frame), most likely due to the highly stochastic nature of SSv2 (random camera jittering, translation, etc). As such, we believe that FVD is a more suitable metric in this scenario since it better measures the realism of dynamic motion in predicted trajectories. We show that our method is able to scale better, comparing POVT to two variants of LVT: LVT-S and LVT-L. LVT-S attends over 8 frames, restricted to similar GLOPs as POVT by limiting $8$ transformer layers. LVT-L attends over 4 frames (same as POVT, excluding object history) with 16 transformer layers, the same as POVT. Therefore, POVT is able to benefit from scaling to larger models under the same compute requirements as LVT while providing similar temporal history. In addition, our model is able to benefit more from bounding box information in complex video when "objectness" is less clear, compared to CATER and MOVi datasets where objects are easily discovered through shape and shading. FitVid does not perform well on this dataset, and collapses into a sequence of blurry frames. MoCoGAN-HD performs the best, able to generate the best samples according FVD. However, it has been demonstrated that GANs tend to perform better on classifier-based metrics such as FVD due to focus on high frequency image statistics whereas VQ-VAE based models throw away through lossy compression~\citep{razavi2019generating}. In addition, we believe that improving other classes of generative models such as likelihood-based models is important, due to ease of training and mode-covering behavior. This can be clearly seen in Figure~\ref{fig:diversity} where GANs collapse to limited futures when conditioned on a single frame.

\textbf{Controllability of Object-centric Models:} Given that POVT adopts a hierarchical framework using bounding boxes, this makes it amenable to certain controllability aspects in generation. Figure~\ref{fig:contr_translate} shows our method able to cleanly translated select objects which account for occlusion.

\begin{table}[]
\begin{adjustwidth}{-2cm}{-2cm}
\begin{tabularx}{\linewidth}{*{2}{>{\centering\arraybackslash}X}}
\caption{Ablation comparing alternative representations for temporal context}
\label{table:abl_bbox}
\begin{tabular}{@{}ccccc@{}}
\toprule
                             & FVD$\downarrow$ & PSNR$\uparrow$ & SSIM$\uparrow$ & LPIPS$\downarrow$ \\ \midrule
No Patch Encoding                   & 267             & 28.5           & 0.890          & 0.0722            \\
Fixed Grid                   & 207             & 28.7           & 0.888          & 0.0770             \\
BBox (ours) & \textbf{168}    & \textbf{30.0}  & \textbf{0.900} & \textbf{0.0598}   \\ \bottomrule
\end{tabular}
&
\caption{Ablation removing each attention component}
\label{table:abl_attn}
\begin{tabular}{@{}lcccc@{}}
\toprule
Method                  & \multicolumn{1}{l}{FVD$\downarrow$} & PSNR$\uparrow$ & SSIM$\uparrow$ & LPIPS$\downarrow$ \\ \midrule
POVT                    & \textbf{164}                        & \textbf{30.1}  & \textbf{0.902} & \textbf{0.0584}   \\
$-$ Base-Base           & 260                                 & 28.1           & 0.881          & 0.0722            \\
$-$ Base-Obj            & 209                                 & 29.6           & 0.896          & 0.0697            \\
$-$ Obj-Obj (over time) & 396                                 & 27.3           & 0.863          & 0.098             \\
$-$ Obj-Obj (per time)  & 174                                 & 29.7           & 0.889          & 0.0643            \\ \bottomrule
\end{tabular}
\end{tabularx}
\end{adjustwidth}
\end{table}

\subsection{Ablations}
\textbf{Object-centric representations provide sufficient temporal history.} For environments where dynamics are primarily composed of objects, we show that object-centric representations are sufficient to represent scene history. Table~\ref{table:abl_cond} shows ablations comparing our method against LVT, with varying number of conditioned frames. Since LVT operates only on full image latents, it benefits greatly from extending the number of conditioned frames further back in time. However, our method, when already conditioned on the history of object-centric representations, shows little to no change in performance when conditioned on more full frames.
\\\\
\textbf{Object-centric representations provide better temporal history of a scene than standard image representations.} We investigate the use of other representations to extend temporal context. In the most general case, we can imagine just encoding previous frames. Table~\ref{table:abl_bbox} shows comparisons of our method which uses bounding box information against generic image latents constructed by encoding fixed patches (similar to ViT) to provide temporal information. At high levels of compression, using bounding box representations is far more beneficial than generic image encodings. In addition, we show that primary benefit of the object representations comes from bounding boxes in combination with region encodings, as opposed to just using bounding boxes.
\\\\
\textbf{Modeling temporal dependencies for objects is crucial for accurate video prediction.} Table~\ref{table:abl_attn} shows the effects of removing each aspect of our object-centric attention mechanism. Removing the object-temporal attention in our models results in the greatest degradation, showing that POVT greatly benefits in modeling longer temporal dynamics through object information.

\section{Discussion}
A primary limitation of our method is that it requires bounding box information, as the benefit the model receives from bounding boxes is restricted by their quality. Bounding boxes also enforce additional restrictions compared to prior methods which only need raw video frames to learn. Nevertheless, we show that there are large gains in computational efficiency when leveraging bounding box information, as even using noisy bounding boxes from a pretrained object detector in Something-Something V2 allows us to scale to larger models, as well produce higher fidelity video samples. Overall, we believe that POVT provides a step in the right direction for a more computationally efficient video generation architecture, a necessity when extending to longer horizon and more complex videos. 

\begin{ack}
The work was in part supported by Panasonic through BAIR Commands, Darpa RACER, and the Hong Kong Centre for Logistics Robotics. In addition, we thank Cirrascale Cloud Services (\href{https://cirrascale.com/}{https://cirrascale.com/}) for providing compute resources.
\end{ack}

\bibliography{neurips_2022}
\bibliographystyle{neurips_2022}

\clearpage
\appendix
\onecolumn
\section{Architecture Hyperparameters}
\label{appendix:arch_hyperparams}
\subsection{VQ-VAE}

\begin{table}[h]
\centering
\caption{VQ-VAE hyperparameters for each dataset}
\begin{tabular}{@{}lccc@{}}
\toprule
                                                           & CATER-CAM / CATER-MOV & MOVi              & Something-Something v2 \\ \midrule
\multicolumn{1}{l|}{Input size}                            & $64 \times 64$        & $64 \times 64$    & $64 \times 64$         \\
\multicolumn{1}{l|}{Latent size}                           & $16 \times 16$        & $16\times 16$     & $16\times16$           \\
\multicolumn{1}{l|}{$\beta$ (commitment loss coefficient)} & 0.25                  & 0.25              & 0.25                   \\
\multicolumn{1}{l|}{Batch size}                            & 128                   & 128               & 128                    \\
\multicolumn{1}{l|}{Learning rate}                         & $7\times 10^{-4}$     & $7\times 10^{-4}$ & $7\times 10^{-4}$      \\
\multicolumn{1}{l|}{Hidden units}                          & 256                   & 256               & 256                    \\
\multicolumn{1}{l|}{Residual units}                        & 128                   & 128               & 128                    \\
\multicolumn{1}{l|}{Residual layers}                       & 2                     & 4                 & 4                      \\
\multicolumn{1}{l|}{Codebook size}                         & 1024                  & 2048              & 4096                   \\
\multicolumn{1}{l|}{Codebook dimension}                    & 128                   & 128               & 256                    \\
\multicolumn{1}{l|}{Encoder filter size}                   & 3                     & 3                 & 3                      \\
\multicolumn{1}{l|}{Upsampling conv filter size}           & 4                     & 4                 & 4                      \\
\multicolumn{1}{l|}{Training steps}                        & 200k                  & 200k              & 200k                  
\end{tabular}
\end{table}

\subsection{Prior Network}
\begin{table}[h]
\centering
\caption{POVT transformer hyperparameters for each dataset}
\begin{tabular}{@{}lccc@{}}
\toprule
                                             & CATER-CAM / CATER-MOV & MOVi                 & Something-Something v2 \\ \midrule
\multicolumn{1}{l|}{Full timesteps}          & 1                     & 1                    & 4                      \\
\multicolumn{1}{l|}{Object timesteps}        & 8                     & 8                    & 8                      \\
\multicolumn{1}{l|}{Object representation size}        & $1\times 1$                     & $1\times 1$                    &   $2\times 2$                    \\
\multicolumn{1}{l|}{Object embedding size}        & 128                     & 128                    & 128                      \\
\multicolumn{1}{l|}{Batch size}              & 32                    & 32                   & 32                     \\
\multicolumn{1}{l|}{Learning rate}           & $1\times 10^{-3}$     & $1\times 10^{-3}$    & $1\times 10^{-3}$      \\
\multicolumn{1}{l|}{Attention heads}         & 4                     & 4                    & 8                      \\
\multicolumn{1}{l|}{Attention Layers}        & 8                     & 8                    & 16                     \\
\multicolumn{1}{l|}{Embedding size}          & 512                   & 512                  & 1024                   \\
\multicolumn{1}{l|}{Feedforward hidden size} & 2048                  & 2048                 & 4096                   \\
\multicolumn{1}{l|}{Dropout}                 & 0.2                   & 0.2                  & 0.2                    \\
\multicolumn{1}{l|}{Attention dropout}       & 0.3                   & 0.3                  & 0.3                    \\
\multicolumn{1}{l|}{Training steps}          & 200k                  & 200k                 & 200k                   \\

                                             & \multicolumn{1}{l}{}  & \multicolumn{1}{l}{} & \multicolumn{1}{l}{}  
\end{tabular}
\end{table}

\section{Dataset Details}
\label{appendix:dataset}
\subsection{CATER}
Since the original CATER dataset does not contain object information, we use a variant of CATER (CATER-CAM) introduced in \citet{kabra2021simone}. CATER-CAM contains around 40k videos with segmentations masks for object. Random camera movement is applied each episode with a maximum of 2 objects moving at a time. Possible object motions include sliding and pick-place actions. We use the default given train-test data split.\\\\
In order to generate CATER-MOV, we modify the original CATER generation codebase (\url{https://github.com/rohitgirdhar/CATER}) to render segmentation masks for each object. We accomplish this by removing all lights and shadows, and performing shadeless rendering for each object set to white and everything else black. Segmentations are then computed through a simple threshold of $0.8$. Rendered videos do not include camera movement, and a maximum of 6 objects move at a time. Possible object motions include sliding, pick-place, and rotation. We generate 10k videos with sequence lengths of 100. We use 10\% of the data as the test set.\\\\
For both datasets, we compute bounding boxes for each object using segmentation masks. If the bounding box area is below a threshold ($5$), we ignore the bounding box. CATER follows the Apache 2.0 license.

\subsection{MOVi}
We generate MOVi using Kubric (\url{https://github.com/google-research/kubric}). We generate 10k trajectories of length 24 with random camera configuration with 5-10 KubricBasic objects. To compute bounding boxes, we follow the same process as CATER using segmentation masks. We use 10\% of the data as the test set. MOVi follows the Apache 2.0 license.

\subsection{Something-Something V2}
Something-Something V2~\cite{goyal2017something} is an action recognition dataset \href{https://developer.qualcomm.com/software/ai-datasets/something-something}{licensed for research use} that contains 220,847 videos. We use a subset of the dataset that has bounding box annotations provided by Something Else~\cite{materzynska2020something}. We follow the original paper and use a simple Kalman Filter-based tracking algorithm~\cite{bewley2016simple} to find bounding box correspondances across frames. In addition, we only train on a subset of data with a maximum of 4 objects. We use the default given train-test split after filtering.

\section{POVT Attention Block Pseudocode}
\label{appendix:code}
\lstset{language=python}
\begin{lstlisting}
def povt_attention_block(base, obj):
    # base: B x T' x H x W x D
    # obj: B x T x K x L x D
    # all attention operations in each block share the same learnable parameters
    
    # Full self-attention over base latent stream
    attn_bb = MaskedMultiheadAttention(base, base)
    
    # Cross attention from object stream to base latent stream
    attn_bo = MultiheadATtention(base, obj)
    
    # Per-object attention over time
    oo_1 = reshape(obj, (B, K, T * L, D))
    attn_oo_1 = MaskedMultiheadAttention(oo_1, oo_1)
    attn_oo_1 = reshape(attn_oo_1, (B, T, K, L, D))
    
    # Per-timestep attention over objects
    oo_2 = reshape(obj, (B, T, K * L, D))
    attn_oo_2 = MaskedMultiheadAttention(oo_2, oo_2)
    attn_oo_2 = reshape(attn_oo_2, (B, T, K, L, D))
    
    # Combine and return
    base = (attn_bb + attn_bo) / 2
    obj = (attn_oo_1 + attn_oo_2) / 2
    
    return base, obj
    
\end{lstlisting}


\end{document}